# Detection of sepsis during emergency department triage using machine learning

Oleksandr Ivanov, PhD – Mednition Inc., Ukraine

Karin Molander, MD, FACEP – Sutter Mills-Peninsula Medical Center, USA

Robert Dunne, MD, FACEP – Ascension Health, USA

Stephen Liu, MD, FACEP – Adventist Health White Memorial, USA

Deena Brecher, MSN, RN, ACNS-BC, CEN, CPEN, FAEN – Mednition Inc., USA

Kevin Masek, MD – San Mateo Medical Center, USA

Erica Lewis, MD – El Camino Hospital, USA

Lisa Wolf, PhD, RN, CEN, FAEN – Emergency Nurses Association, USA

Debbie Travers, PhD, RN, FAEN – Duke University, USA

Deb Delaney, DNP, MHA, RN, CEN – Mednition Inc., USA

Kyla Montgomery – Mednition Inc., USA

Christian Reilly – Mednition Inc., USA

Corresponding author: Christian Reilly, creilly@mednition.com

## Abstract

Sepsis is a life-threatening condition with organ dysfunction and is a leading cause of death and critical illness worldwide. Even a few hours of delay in the treatment of sepsis results in increased mortality. Early detection of sepsis during emergency department triage would allow early initiation of lab analysis, antibiotic administration, and other sepsis treatment protocols. The purpose of this study was to compare sepsis detection performance at ED triage (prior to the use of laboratory diagnostics) of the standard sepsis screening algorithm (SIRS with source of infection) and a machine learning algorithm trained on EHR triage data. A machine learning model (KATE Sepsis) was developed using patient encounters with triage data from 16 participating hospitals. KATE Sepsis and standard screening were retrospectively evaluated on the adult population of 512,949 medical records. KATE Sepsis demonstrates an AUC of 0.9423 (0.9401 - 0.9441) with sensitivity of 71.09% (70.12% - 71.98%) and specificity of 94.81% (94.75% - 94.87%). Standard screening demonstrates an AUC of 0.6826 (0.6774 - 0.6878) with sensitivity of 40.8% (39.71% - 41.86%) and specificity of 95.72% (95.68% - 95.78%). The KATE Sepsis model trained to detect sepsis demonstrates 77.67% (75.78% - 79.42%) sensitivity in detecting severe sepsis and 86.95% (84.2% - 88.81%) sensitivity in detecting septic shock. The standard screening protocol demonstrates 43.06% (41% - 45.87%) sensitivity in detecting severe sepsis and 40% (36.55% - 43.26%) sensitivity in detecting septic shock. Future research should focus on the prospective impact of KATE Sepsis on administration of antibiotics, readmission rate, morbidity and mortality.

## 1. Introduction

Sepsis is a life-threatening condition with organ dysfunction caused by a dysregulated host response to infection (Singer et al. 2016) and is a leading cause of death and critical illness worldwide (Angus et al. 2001, Vincent et al. 2014, Fleischmann et al. 2015, Liu et al. 2014, Fleischmann-Struzek et al. 2020, Vincent et al. 2020).

The global incidence of sepsis is estimated at 32 million with 5.3 million deaths per year (Fleischmann, 2016). In the United States alone more than 1.7 million patients are diagnosed with sepsis annually and nearly 270,000 die from sepsis (CDC, 2021). In 2017, the United States spent over $38 billion on sepsis treatment, making it the most expensive condition to treat (Liang, 2020). Patients who survive sepsis often have long-term health and social consequences (Iwashyna et al. 2010). Despite advances in medical treatment, the reported incidence of sepsis has failed to decrease (Iwashyna et al. 2012, Gaieski et al. 2013, Rhee et al. 2017), and sepsis remains the primary cause of death from infection (Singer et al. 2016).

The difficulty of treating sepsis arises from challenges in early detection, with subtle and heterogeneous symptoms at the onset of the disease and often quick progression to life-threatening stages of severe sepsis and septic shock (Vincent, 2016, de Grooth et al. 2018). Even a few hours of delay in the treatment of sepsis results in increased mortality (Seymour et al. 2017, Liu et al. 2017, Ferrer 2014, Kumar et al. 2006). Early detection of sepsis allows for close monitoring, early testing and treatment, and has been shown to reduce mortality, morbidity and cost of care (Rivers et al. 2001, Nguyen et al. 2007, Seymour et al. 2017). The Surviving Sepsis Campaign International Guidelines recommend that administration of IV antimicrobials be initiated immediately, ideally within 1 hour of recognition for both sepsis and septic shock onset (Evans et al. 2021). Early and accurate detection of sepsis is critical to reduce the negative impacts of sepsis on the population.

Despite high prevalence, early and accurate detection of sepsis remains elusive. Existing rule-based methods for early detection of sepsis, such as systemic inflammatory response syndrome (SIRS) criteria and quick sequential organ failure assessment score (qSOFA) have low sensitivity and specificity (Islam et al. 2019, Filbin et al. 2018) and do not show significant change in clinical outcomes (Nelson et al. 2011, Hooper et al. 2012, Nguyen et al. 2014).

Machine learning (ML) has been studied as a promising tool to detect sepsis (Schinkel et al. 2019, Islam et al. 2019, Fleuren et al. 2020, Teng et al. 2020, Giacobbe et al. 2021). The majority of research is focused on applying ML to predict sepsis within hospital intensive care units (ICU). Some studies used ML to predict sepsis for patients throughout their stay in the ED using vital signs and lab results as model features (Haug and Ferraro, 2016**,** Delahanty et al. 2019) for severe sepsis or septic shock (Jones et al. 2012, Brown et al. 2016, Nachimuthu and Haug, 2012, McCoy and Das, 2017). Sepsis, or the infection causing sepsis, starts outside of the hospital in nearly 87% of cases (CDC, 2021). Ideally sepsis should be detected when a patient arrives at an Emergency Department (ED) during triage. Accurate detection of sepsis during the triage assessment would allow early initiation of lab analysis, antibiotic administration, and other sepsis treatment protocols. The vast amount of extant literature for ML in sepsis focuses on ICU patients, and to the best of our knowledge, there are no publications with ML models that focus on detection of sepsis at ED triage.

The goal of this study was to develop an accurate ML model, which we call "KATE Sepsis", to detect sepsis with the limited information available at ED triage, where no lab results are present, and compare its performance to a standard screening protocol (SIRS with documented source of infection), which was used at all study sites.

## 2. Methods

### 2.1. Data collection

Human subjects protection and conflicts of interest: Prior to data collection, IRB exemption was obtained from the WCG IRB (IRB registration number: 13574120). All protected health information (PHI) was redacted from the datasets, in accordance with 45 CFR § 164.514 for de-identifying PHI to safe-harbor standards. De-identified raw text files were mapped and consolidated to a multi-hospital data model, preserving the differences between sites.

## 2.2. Study Sites and demographics

We conducted a retrospective study using ED triage data from 16 hospital sites (Site 1 through 16) across California, Oregon, and Hawaii from Feb 2015 to July 2021, 615,581 medical records with 9,336 sepsis diagnoses. In this study we used only the adult population (18 years or older) since sepsis in pediatric patients is very rare (79 sepsis incidents for 99,043 pediatric records). We also removed 3589 medical records missing four or more vital signs (pulse rate, respiratory rate, temperature, oxygen saturation, systolic and diastolic blood pressure). Demographics and sepsis, severe sepsis and septic shock prevalence for the remaining adult population of 512,949 medical records from 16 study sites are presented in Table 1.

**Table 1. Demographic data and sepsis diagnosis for 16 study hospital sites adult population of 512,949 medical records.**

| Hospital site | Dates | Number of records | Males | Females | All Sepsis diagnosis | Severe sepsis | Septic shock |
|---|---|---|---|---|---|---|---|
| Site 1 | Feb 2015 - May 2018 | 131054 | 54689 (41.73%) | 76352 (58.26%) | 2807 (2.14%) | 574 (0.44%) | 152 (0.12%) |
| Site 2 | Jan 2020 - Dec 2020 | 53796 | 24983 (46.44%) | 28812 (53.56%) | 1094 (2.03%) | 172 (0.32%) | 109 (0.2%) |
| Site 3 | Jul 2020 - Jun 2021 | 40165 | 17593 (43.8%) | 22570 (56.19%) | 402 (1.0%) | 63 (0.16%) | 41 (0.1%) |
| Site 4 | Jan 2020 - Dec 2020 | 51487 | 24180 (46.96%) | 27301 (53.03%) | 1173 (2.28%) | 372 (0.72%) | 138 (0.27%) |
| Site 5 | Oct 2019 - Oct 2020 | 41075 | 19072 (46.43%) | 22002 (53.57%) | 1072 (2.61%) | 124 (0.3%) | 91 (0.22%) |
| Site 6 | Jul 2020 - Jul 2021 | 23037 | 9200 (39.94%) | 13837 (60.06%) | 54 (0.23%) | 10 (0.04%) | 4 (0.02%) |
| Site 7 | Jan 2020 - Dec 2020 | 17457 | 8376 (47.98%) | 9080 (52.01%) | 213 (1.22%) | 23 (0.13%) | 12 (0.07%) |
| Site 8 | Jul 2020 - Jun 2021 | 13326 | 6448 (48.39%) | 6874 (51.58%) | 158 (1.19%) | 5 (0.04%) | 7 (0.05%) |
| Site 9 | Jul 2020 - | 27600 | 11746 | 15853 | 141 | 20 | 11 |

**Table 1. Demographic data and sepsis diagnosis for 16 study hospital sites adult population of 512,949 medical records.**

| | | | | | | | |
|---|---|---|---|---|---|---|---|
| | Jul 2021 | | (42.56%) | (57.44%) | (0.51%) | (0.07%) | (0.04%) |
| Site 10 | Oct 2019 - Sep 2020 | 27074 | 12311 (45.47%) | 14762 (54.52%) | 679 (2.51%) | 39 (0.14%) | 32 (0.12%) |
| Site 11 | Jan 2020 - Dec 2020 | 12350 | 5807 (47.02%) | 6540 (52.96%) | 102 (0.83%) | 5 (0.04%) | 9 (0.07%) |
| Site 12 | Jul 2020 - Jun 2021 | 34273 | 16859 (49.19%) | 17408 (50.79%) | 642 (1.87%) | 152 (0.44%) | 47 (0.14%) |
| Site 13 | Jul 2020 - Jun 2021 | 23547 | 11696 (49.67%) | 11851 (50.33%) | 474 (2.01%) | 83 (0.35%) | 39 (0.17%) |
| Site 14 | Jul 2020 - Jun 2021 | 7612 | 3757 (49.36%) | 3846 (50.53%) | 85 (1.12%) | 9 (0.12%) | 8 (0.11%) |
| Site 15 | Jan 2020 - Dec 2020 | 5994 | 2896 (48.31%) | 3097 (51.67%) | 102 (1.7%) | 19 (0.32%) | 5 (0.08%) |
| Site 16 | Jul 2020 - Jun 2021 | 3102 | 1501 (48.39%) | 1598 (51.52%) | 59 (1.9%) | 9 (0.29%) | 0 (0%) |
| All sites | Feb 2015 - Jul 2021 | 512949 | 231114 (45.06%) | 281783 (54.93%) | 9257 (1.8%) | 1679 (0.33%) | 705 (0.14%) |

### 2.3. Sepsis diagnosis definition

We used sepsis ED diagnosis as a positive label in this study. Sepsis diagnoses in the ED were assigned within a 24 hour period after patient arrival. Sepsis diagnoses were assigned by ED clinicians based on Sepsis-2 definition, which was used at the study sites for Centers for Medicare and Medicaid Services (CMS) for compliance reporting. We considered sepsis, severe sepsis, and septic shock diagnoses as a “sepsis” diagnosis for the KATE Sepsis model. We also considered sepsis as SIRS with a documented diagnosis of infection.

Diagnoses were written as free text from ED provider notes. We extracted sepsis from the free text ED diagnoses using Clinical terms extraction algorithm, which is described in Section 2.5.2.

## 2.4. Sepsis screening protocol

We also evaluated a common rule-based sepsis screening protocol, SIRS with source of infection, that was an instituted sepsis screening protocol at each study site. *SIRS* protocol is defined as two or more SIRS vitals outside normal range: temperature < 36 or > 38 °C, pulse rate > 90 or respiratory rate > 20 (Bone et al. 1992). The SIRS protocol also includes white blood cell count, but we omitted this because labs are not available during ED triage. *Standard screening protocol* is defined as two or more SIRS vitals outside normal range and a source of infection documented anywhere in the triage assessment (Bone et al. 1992). See Supplementary Table 1 for infection types. All study sites prior to the research had adopted the standard screening protocol at the time of ED Triage. Results for the Standard screening protocol are presented in Section 3 and compared to the KATE Sepsis model.

## 2.5. Machine Learning

### 2.5.1. Features overview

In this study, numerical, categorical, and free text data from the EHR were used. Numerical features are represented by age, triage vital signs (pulse rate, respiratory rate, temperature, systolic blood pressure, diastolic blood pressure, and oxygen saturation), primary pain intensity, Glasgow Coma Scale score and point of care (POC) blood glucose value. Numerical data were transformed into features after removing physiologically impossible outliers. Categorical data are represented by sex, arrival mode (walk-in, ambulance, wheelchair, ambulatory, not ambulatory, stretcher or carried, law enforcement) and immunization status (value 1 if immunizations are up to date, 0 if not up to date). Post-triage data, such as labs or vitals signs after triage, were not used as model features. Clinical terms were extracted from chief complaints, medical history, family history, surgeries, triage treatment and prior to arrival medications using Clinical concepts extraction algorithm (Section 2.5.2).

### 2.5.2. Clinical concepts extraction algorithm

Accurate extraction of clinical terms from patient record free text is a prerequisite to form a complete understanding of each patient and can enhance ML-based clinical decision models. This has been a primary challenge in building ML-based clinical decision support tools for clinicians which leverage clinical raw text evidence. We developed the Clinical concepts extraction algorithm to accurately extract medical terms from free text (Ivanov et al. 2020).

In the Clinical concepts extraction algorithm we use the following steps to process raw text:

1. Sentences tokenization
2. Words tokenization
3. Text normalization
4. Part of speech tagging
5. Chunking
6. Extraction of clinical terms

Steps 1-5 are done with the OpenNLP Java library. Extraction of clinical terms in step 6 is done with following steps:

1. Noun phrases (NP) and verb phrases (VP) are extracted from chunker (step 5)
2. Text in each NP is permuted in all combinations for all phrases
3. All text combinations are matched against the UMLS database (Bodenreider, 2004), clinical terms are extracted based on matching to UMLS terms
4. For each medical term unique UMLS code (CUI) is extracted and used as a feature.

The breakdown of unique clinical concepts extracted from raw text from different sections of medical records are presented in Table 2.

| Table 2. Count of unique and total number of clinical concepts by section of medical record extracted using clinical concepts extraction algorithm from the adult population of 512,949 medical records from 16 study sites. | | | |
|---|---|---|---|
| **Section** | **Number of unique clinical concepts** | **Total number of clinical concepts** | **Mean number of clinical concepts per record** |
| **Reason for visit** | 10033 | 3892085 | 7.59 |
| **Medical history** | 15503 | 3075296 | 6 |
| **Family history** | 912 | 162524 | 0.32 |
| **Surgeries** | 6337 | 854771 | 1.67 |
| **Prior to arrival medications** | 461 | 73653 | 0.14 |
| **Triage treatment** | 286 | 97238 | 0.19 |

### 2.5.3. Machine learning algorithms

We used XGBoost and logistic regression stack as an ML algorithm in this study. XGBoost is a method from the gradient tree boosting family. Gradient boosting is a method of sequential building of decision trees in which each subsequent tree is built on the subset of data where previous trees made the most mistakes in classification (Chen, 2016). XGBoost was designed to have an efficient model training performance for large sparse datasets (Chen, 2016) which is common for clinical data. Predicted probability for each medical record from XGBoost was used as a feature for logistic regression. We assigned the parameter "class weight" to "balanced" to overcome the problem of class imbalance for sepsis (low percent of positive sepsis diagnosis). Features missing values were handled by XGBoost by default. Missing values KATE Sepsis was trained and evaluated using 5-fold cross-validation on 512,949 adult medical records from 16 study sites.

### 2.6. Software

Java 8.0 and OpenNLP Java library were used to develop Clinical concepts extraction. Python 3.8 was used for ML pipeline development. XGBoost 1.4 library was used to build KATE Sepsis. Sklearn and Scipy libraries were used for model evaluation and statistical analysis.

### 2.7. Data availability

The medical records used in this study are not publicly available due to the Business Associate Agreement with the participating hospital sites. A small sample of de-identified records is available from the corresponding author upon reasonable request.

## 3. Results

### 3.1. Primary results

Results on 5-fold cross-validation for KATE Sepsis and standard screening are presented in Table 3. KATE Sepsis demonstrates significantly higher performance compared to the standard screening protocol. Specifically, KATE Sepsis demonstrates an AUC of 0.9423 (0.9401 - 0.9441) with sensitivity of 71.09% (70.12% - 71.98%) and 94.81% (94.75% - 94.87%). Standard screening demonstrates an AUC of 0.6826 (0.6774 - 0.6878) with sensitivity of 40.8% (39.71% - 41.86%) and specificity of 95.72% (95.68% - 95.78%).

**Table 3.** Performance metrics and 95% confidence intervals (in parentheses) for the adult population from 16 study hospital sites (Feb 2015 to Jul 2021) of KATE Sepsis and standard screening evaluated 512,949 medical records with 9,257 sepsis diagnoses using 5-fold cross validation.

| Group | AUC | Sensitivity | Specificity | F1-score | Accuracy | Precision |
|---|---|---|---|---|---|---|
| KATE Sepsis | 0.9423 (0.9401 - 0.9441) | 0.7109 (0.7012 - 0.7198) | 0.9481 (0.9475 - 0.9487) | 0.31365 (0.3068 - 0.3193) | 0.94385 (0.9432 - 0.9444) | 0.20121 (0.1961 - 0.2055) |
| Standard screening | 0.6826 (0.6774 - 0.6878) | 0.408 (0.3971 - 0.4186) | 0.9572 (0.9568 - 0.9578) | 0.21844 (0.2128 - 0.2247) | 0.94731 (0.9468 - 0.9479) | 0.14915 (0.1448 - 0.1539) |

The KATE Sepsis model trained to detect sepsis demonstrates 77.67% (75.78% - 79.42%) sensitivity in detecting severe sepsis and 86.95% (84.2% - 88.81%) sensitivity in detecting septic shock. In contrast, the standard screening protocol demonstrates 43.06% (41% - 45.87%) sensitivity in detecting severe sepsis and 40% (36.55% - 43.26%) sensitivity in detecting septic shock.

We also compared the percentage of correct sepsis diagnosis predictions of standard screening versus KATE Sepsis. KATE Sepsis detects 92.03% of sepsis diagnoses that the standard screening protocol predicts correctly, whereas the standard screening predicts correctly only 52.82% of sepsis cases that KATE Sepsis predicts correctly.

Feature importances are presented in Figure 1. Feature importances demonstrate that other features besides the standard screening features (temperature, pulse rate and respiratory rate), such as age, blood pressure, weight, diabetes, lassitude and others, play a role in sepsis detection.

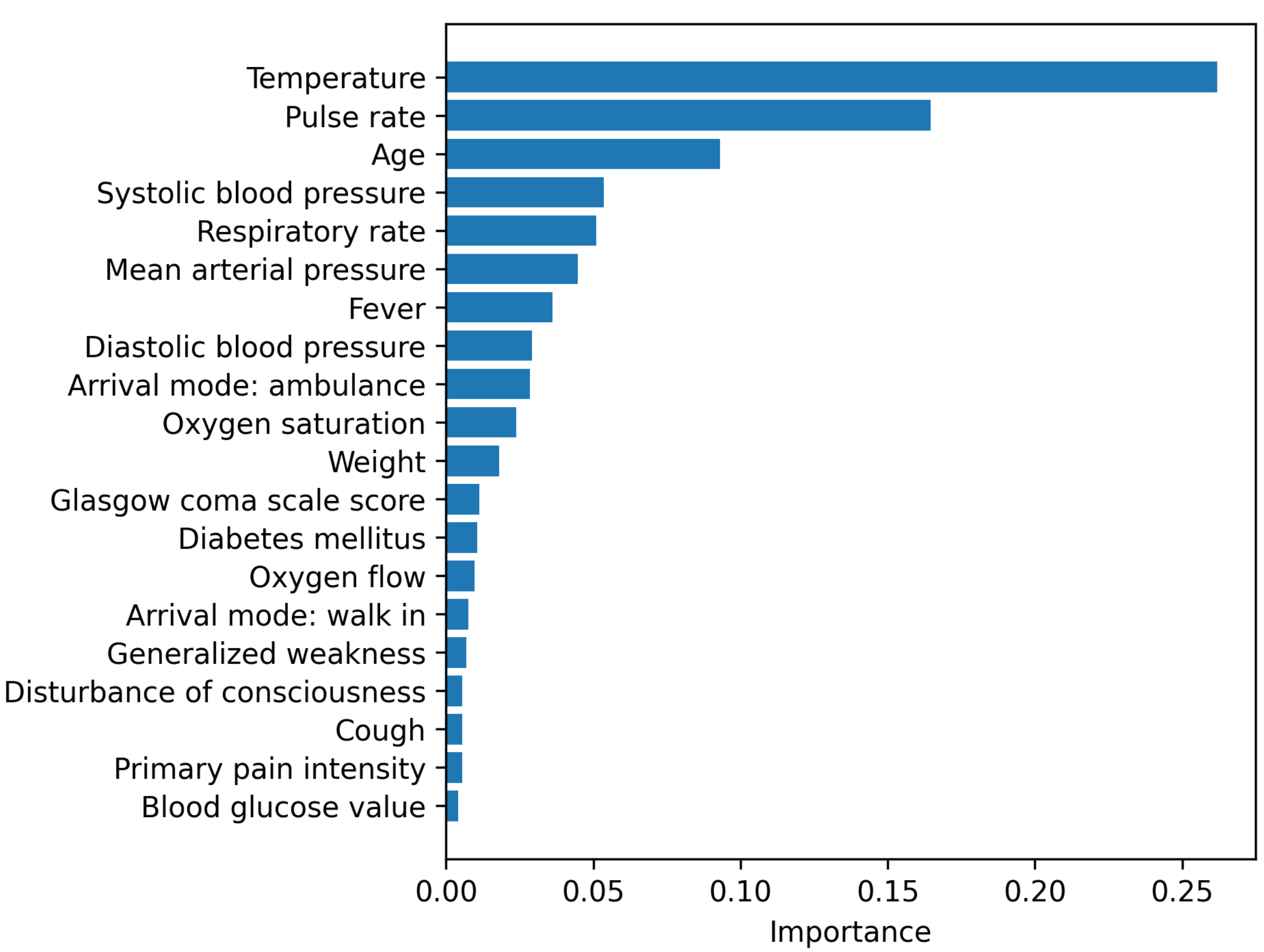

**Figure 1.** Feature importances of the KATE Sepsis model for the top-20 features ranked by their relative importance.

We added distributions for sepsis diagnosis, sepsis predictions for the standard screening protocol and KATE sepsis by sex in Supplementary tables 2-4.

### 3.2. Compare of KATE Sepsis and standard screening

KATE Sepsis has a similar specificity to standard screening but significantly outperforms in sensitivity (see Section 3.1.). The reason the standard screening protocol demonstrates low performance in detecting sepsis at triage is that the abnormal values of vital signs that are used in SIRS (temperature, pulse rate and respiratory rate) are often not sufficient to detect sepsis at triage. This is illustrated in Figure 2, which illustrates distributions of SIRS vital signs for patients without sepsis and for patients with sepsis. The red dashed lines in Figure 2 indicate abnormal values of triage vital signs above or below which SIRS protocol is triggered (see Section 2.6). Figure 2 demonstrates that although patients with sepsis more often present to ED with abnormal SIRS vital signs compared to patients without sepsis, the majority of patients diagnosed with sepsis have triage vital signs that do not trigger SIRS protocol.

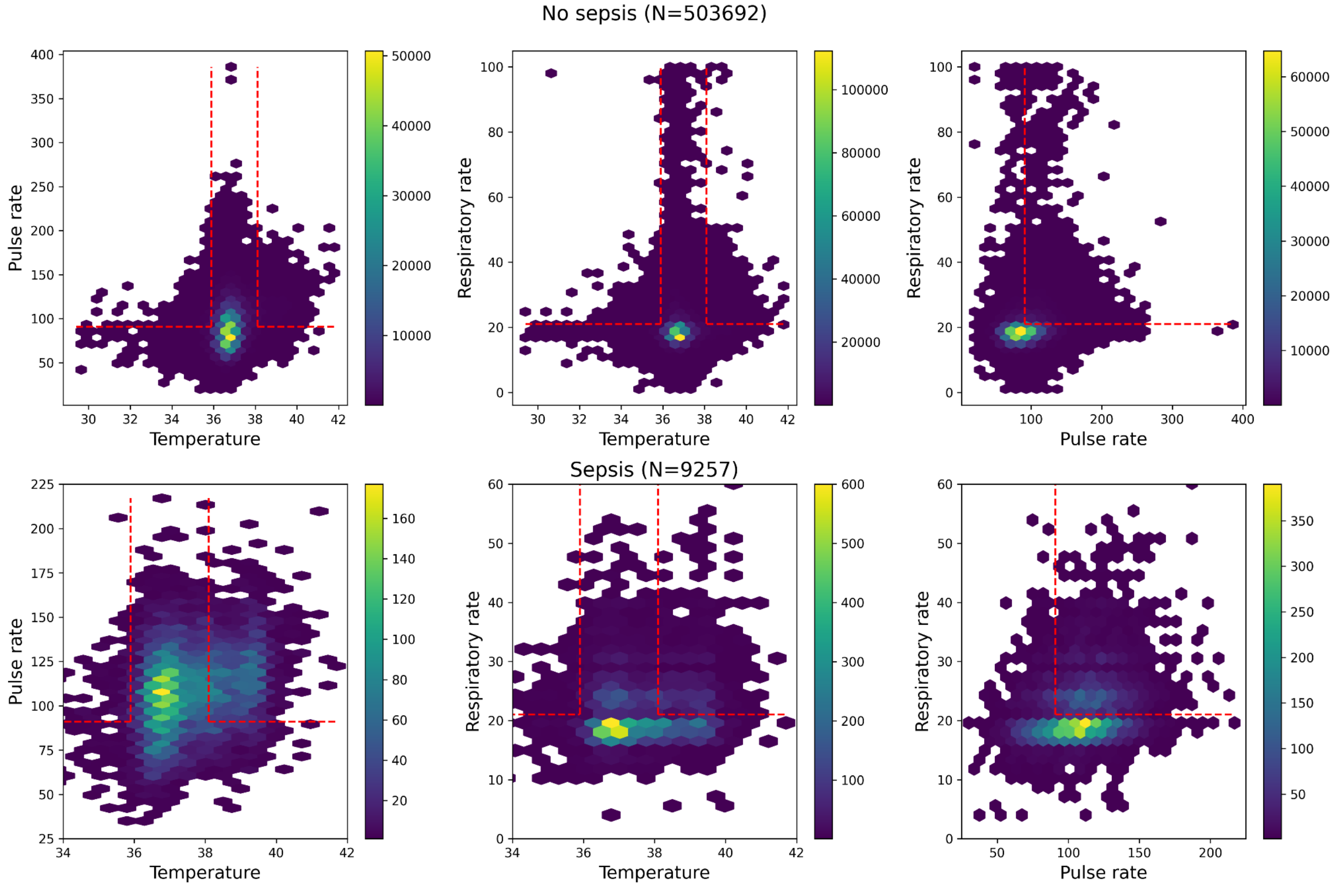

**Figure 2.** Heatmaps of SIRS vital signs (temperature, pulse rate and respiratory rate) for adult patients (N=512,954) for 16 study sites without sepsis (top panel, N=503,692) and with sepsis diagnosis (bottom panel, N=9,257). The color of each hexagon indicates the number of medical records with given values of vital signs at triage. The red dotted lines indicate rectangular areas for vital signs that trigger SIRS. The majority of patients diagnosed with sepsis on the sepsis panel do not fall into the area constrained by SIRS vital signs.

## 4. Discussion

Sepsis is a leading cause of death in hospitals (Liu et al. 2014), hospital readmissions (Fingar and Washington, 2015), and cost of care (Buchman et al. 2020). Early and accurate screening is critical for rapid treatment, potentially improving the patient's outcomes, and reducing mortality risk (Rivers et al. 2001, Nguyen et al.

2007, Seymour et al. 2017). Nurses at triage have an important role in recognizing high risk patients and prioritizing patients for rapid intervention.

The purpose of this study was to compare sepsis detection performance at ED triage (prior to the use of laboratory diagnostics) of the standard sepsis screening protocol (SIRS with source of infection) and a machine learning algorithm trained on EHR triage data (KATE Sepsis).

This study demonstrated that an ML approach can accurately detect sepsis diagnosis with data available at triage, and prior to diagnostic results such as labs. Standard sepsis screening protocol demonstrates low performance in detecting sepsis at triage. Standard screening underperforms at triage because septic patients often do not present with temperature, pulse rate and respiratory rate that trigger SIRS protocol (Section 3.2). In contrast to standard screening, KATE Sepsis uses other features, such as age, blood pressure, chief complaints, medical history and others, which allows higher performance in sepsis detection (Figure 1). Detection of patients with severe sepsis or septic shock at triage is crucial, because such patients are at high risk of morbidity and mortality (Schoenberg et al. 1998, Bauer et al. 2020). The KATE Sepsis model detects sepsis with high accuracy, especially for patients diagnosed in the ED with severe sepsis and septic shock (Section 3.1). Standard screening demonstrates low performance for severe sepsis and septic shock detection.

The lack of predictive power of the standard screening protocol results in significant underdetection of patients with sepsis at triage and can lead to delays in treatments and potential deterioration of patient's conditions. Machine learning algorithms, such as KATE Sepsis, can significantly increase detection rate of patients with sepsis at triage and therefore demonstrate promise to reduce time to treatment, morbidity and mortality. KATE Sepsis can be used at ED for diagnostic screening of sepsis at ED triage, predicting risk of sepsis for further evaluation by clinicians.

KATE Sepsis provides 77.3% higher sensitivity compared to standard screening, with a similar specificity, a follow up question is whether standard screening can be replaced with KATE Sepsis. Although KATE Sepsis cannot fully replace standard screening protocol, it detects the majority of all sepsis diagnoses that standard screening detects. Standard screening true positive cases are detected 92.03% of the time by the KATE Sepsis model (Section 3.1).

The definition of sepsis changes over time (Gül et al. 2017). Sepsis diagnoses were assigned by ED clinicians based on Sepsis-2 definition, which was used at the study sites for Centers for Medicare and Medicaid Services (CMS) for compliance reporting. We used sepsis ED diagnosis based as true labels. In 2016 a new definition of sepsis, Sepsis-3, was introduced based on Sequential [Sepsis-related] Organ Failure Assessment (SOFA) (Singer et al. 2016). The lack of a gold standard definition of sepsis implemented in hospitals creates a problem both for clinicians and researchers. The discrepancy of ED physician sepsis diagnosis with Sepsis-2 and Sepsis-3 formal criteria is not known and should be a topic of a follow up research. Future research should focus on training and evaluation of ML models using Sepsis-3 as a true label and measuring clinical outcomes of such models.

During the course of study, outcomes data: such as antibiotics administration, length of stay, mortality, readmission rate was not available. Future prospective research will focus on the impact of KATE Sepsis and other screening methods on clinical decision making and patient outcomes, including time to antibiotics administration, readmission rate and mortality.

In this study we used information available at triage, i.e. without labs. Future research will study how the availability of data during patient stay at ED can improve sepsis detection.

Future research should also focus on prospective analysis of KATE Sepsis live performance as a diagnostic screening tool at ED during triage.

## Supplementary data

| **Supplementary Table 1.** Example potential infection signs and symptoms for various systems used in the standard screening algorithm | |
|---|---|
| **System** | **Examples** |
| Respiratory | Cough, pneumonia |
| GI/GU | UTI, gastroenteritis |
| Skin | Cellulitis, abscess, petechiae |
| Neurological | Nuchal rigidity, photophobia, Altered level of consciousness with fever |
| Oral Maxillofacial | Facial swelling, periorbital cellulitis |
| Oncological | Neutropenia with fever, central line infection |
| Other | Fever status post surgery, postpartum |

| **Supplementary Table 2.** Distribution of sepsis diagnosis by sex for the adult population of 512,949 medical records from 16 study hospitals. | | |
|---|---|---|
| **Sex** | **No sepsis** | **Sepsis** |
| **Female** | 277284 (54.06%) | 4499 (0.88%) |
| **Male** | 226356 (44.13%) | 4758 (0.93%) |

| **Supplementary Table 3.** Distribution of sepsis diagnosis predictions at triage by the standard screening protocol by sex for the adult population of 512,949 medical records from 16 study hospital sites. | | |
|---|---|---|
| **Sex** | **No sepsis** | **Sepsis** |
| **Female** | 268748 (52.4%) | 13035 (2.54%) |
| **Male** | 218826 (42.66%) | 12288 (2.4%) |

| **Supplementary Table 4.** Distribution of sepsis diagnosis predictions at triage by the KATE sepsis model by sex for the adult population of 512,949 medical records from 16 study hospital sites. | | |
|---|---|---|
| **Sex** | **No sepsis** | **Sepsis** |
| **Female** | 265397 (51.74%) | 16386 (3.19%) |
| **Male** | 214796 (41.88%) | 16318 (3.18%) |